\documentclass[final]{template}

\usepackage{times}
\usepackage{epsfig}
\usepackage{graphicx}
\usepackage{amsmath}
\usepackage{amssymb}
\usepackage{mathtools}
\usepackage{url}
\usepackage{color}
\usepackage{float}
\usepackage{array}
\usepackage{multirow}
\usepackage{arydshln}
\usepackage{longtable}

\usepackage{multicol}
\usepackage{lipsum}
\usepackage{adjustbox}
\usepackage[pagebackref=true,breaklinks=true,colorlinks,bookmarks=false]{hyperref}

\begin{document}

\title{Monocular Depth Estimation Primed by Salient Point Detection and Normalized Hessian Loss}

\author{Lam Huynh$^1$ \qquad
Matteo Pedone$^1$ \qquad
Phong Nguyen$^1$ \\
Jiri Matas$^2$ \qquad
Esa Rahtu$^3$ \qquad
Janne Heikkil\"a$^1$ \\
\small{$^1$Center for Machine Vision and Signal Analysis, University of Oulu} \\ 
\small{$^2$ Center for Machine Perception, Czech Technical University, Czech Republic} \\ 
\small{$^3$Computer Vision Group, Tampere University} }

\maketitle

\begin{abstract}
Deep neural networks have recently thrived on single image depth estimation. That being said, current developments on this topic highlight an apparent compromise between accuracy and network size. This work proposes an accurate and lightweight framework for monocular depth estimation based on a self-attention mechanism stemming from salient point detection. Specifically, we utilize a sparse set of keypoints to train a FuSaNet model that consists of two major components: Fusion-Net and Saliency-Net. In addition, we introduce a normalized Hessian loss term invariant to scaling and shear along the depth direction, which is shown to substantially improve the accuracy. The proposed method achieves state-of-the-art results on NYU-Depth-v2 and KITTI while using 3.1-38.4 times smaller model in terms of the number of parameters than baseline approaches. Experiments on the SUN-RGBD further demonstrate the generalizability of the proposed method.
\end{abstract}

\section{Introduction}

Acquiring accurate depth information from 2D images is crucial for computer vision, ranging from robotics, scene understanding, and augmented reality. Traditional multi-view setups~\cite{hartley2003multiple,szeliski2011structure} obtain accurate results, but measurements are sparse and heavily depend on feature extractions, while active depth sensors are costly. On the other hand, recent learning-based monocular depth estimation approaches~\cite{chen2019structure,facil2019cam,Hu2018RevisitingSI,huynh2020guiding,lee2019big,liu2019planercnn,liu2018planenet,ramamonjisoa2019sharpnet,Ranftl2021,yang2021transformers} have achieved promising results making them potential alternatives to conventional multi-view methods.

Learning-based monocular depth estimation relies on the idea of training a model to predict dense depth values for every RGB pixel. However, training such networks typically requires substantial amount of data and massive network architectures. State-of-the-art methods tend to employ large encoders like ResNet~\cite{fu2018deep,laina2016deeper,qi2018geonet,ramamonjisoa2019sharpnet}, ResNext-101~\cite{Yin2019enforcing}, SeNet-154~\cite{chen2019structure,Hu2018RevisitingSI}, Transformer~\cite{Ranftl2021,yang2021transformers}, with sophisticated decoder strategies~\cite{chen2019structure,Hu2018RevisitingSI,laina2016deeper}, and train with huge dataset such as PBRS~\cite{zhang2017physically}, \textit{MIX 6}~\cite{Ranftl2021} to achieve high accuracy. On the contrary, fast solutions~\cite{ignatov2021fast,wofk2019fastdepth} suffer from low precision, manifesting an apparent compromise between accuracy and network size.

Depth completion is a related problem where the aim is to densify a sparse depth map by using machine learning techniques. The sparse depth measurements allow regularizing the nearby depth values, and as a consequence, high accuracy can be achieved with considerably smaller networks~\cite{hu2020PENet,huynh2020boosting,park2020non,zhang2018deep}. However, the obvious disadvantage of depth completion is the requirement of additional data that is often obtained with active sensors such as LiDARs or ToF cameras.

In this paper, we propose an approach where known depth measurements are replaced with salient points to regularize the depth map. A major benefit compared to depth completion is that these salient points are determined from monocular RGB images while still providing similar advantages as the additional depth data. In this context, salient points are assumed to be image details where the local structure reveals the depth accurately even from a single view. Thus, it can be also considered to be a self-attention mechanism.
For this purpose, we first train confidence predictors to highlight important keypoint positions from an RGB image as a confidence map. This map is then used to generate salient points where predicted depth values tend to be more accurate. These points are utilized to enhance the performance of our network similar to depth completion. Moreover, to further assist the network in learning local structures from a single image, we also introduce a normalized Hessian loss term invariant to scaling and shear along the depth direction. In summary, our work makes the following contributions:

\begin{figure*}[!t]
\begin{center}
\includegraphics[width=0.99\linewidth]{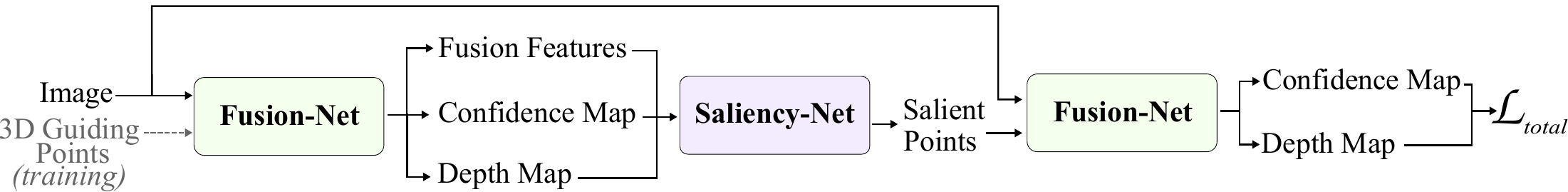}
\end{center}
  
  \caption{The overall structure of FuSaNet. At training time, the 3D guiding points are sampled from the ground truth depth map at keypoints locations. The Fusion-Net first takes the RGB image and 3D guiding points as inputs to produce a confidence map, a depth map, and a fusion features tensor. These outputs are used to detect the salient points that are fed to the Fusion-Net to generate the final confidence and depth map. At inference time, only the RGB image is required as input to the framework.}
\label{fig:overview}  \vspace{-0.45cm}
\end{figure*}

\begin{itemize}
 \item We propose a novel FuSaNet architecture for monocular depth estimation that utilizes a self-attention mechanism based on salient point detection and 3D feature fusion to improve the depth estimation accuracy.
 \item We utilize a normalized Hessian loss term that is insensitive to the generalized bas-relief (GBR) ambiguity.
 \item We achieve state-of-the-art results with indoor (NYU-Depth-v2, SUN-RGBD) and outdoor (KITTI) scenes while using significantly fewer parameters than baseline approaches.
\end{itemize}

\noindent Implemention of FuSaNet will be made publicly available upon publication of the paper.

\section{Related work} \label{sec:related_work}
\noindent \textbf{Monocular depth estimation:}
The interest in single image depth estimation has dramatically increased over the recent years since first proposed by Saxena et al.~\cite{saxena2006learning} and Eigen et al.~\cite{eigen2015predicting,eigen2014depth}. Pioneering studies obtained accurate depth maps mainly by using large network architectures~\cite{chen2019structure,Hu2018RevisitingSI,laina2016deeper}. Then, Jiao et al.~\cite{jiao2018look} exploited semantic information while Qi et al.~\cite{qi2018geonet} utilized the duality between depth and surface normal to improve accuracy. Fu et al.~\cite{fu2018deep} re-defined monocular depth estimation as a classification problem that later on inspired Ren et al.~\cite{ren2019deep} to build their work as a mixture of both tasks. Reflecting the natural scale ambiguity, Lee et al.~\cite{lee2019monocular} suggested estimating relative instead of absolute depth values while Facil et al.~\cite{facil2019cam} trained a network to recognize the camera calibration models. Recent depth estimation methods focused on learning the monocular priors such as occlusion~\cite{ramamonjisoa2019sharpnet}, planarity both explicitly~\cite{liu2019planercnn,liu2018planenet,Yin2019enforcing} and implicitly~\cite{huynh2020guiding,lee2019big}. Gonzalez and Kim~\cite{gonzalezbello2020forget} proposed to synthesize the right view from the left view for training from stereo images. Yang et al.~\cite{yang2021transformers} and Ranftl et al.~\cite{Ranftl2021} utilize transformer modules to estimate high-quality depth maps, while in contrast~\cite{ignatov2021fast,wofk2019fastdepth} proposed fast depth estimation methods. However, there is a clear trade-off between accuracy and model size.

\noindent \textbf{Depth completion:}
Depth completion methods produce a dense depth map starting from a set of incomplete depth measurements. Pioneering works from Diebel and Thrun~\cite{diebel2006application} and Hawe et al.~\cite{hawe2011dense} proposed using Markov Random Fields or Wavelet analysis to tackle this problem. Then later, Uhrig et al.~\cite{uhrig2017sparsity} utilized sparse convolution to build a network taking into account different input sparsities. Jaritz et al.~\cite{jaritz2018sparse} used semantic segmentation while Ma et al.~\cite{mal2018sparse} directly concatenated the sparse depth with the RGB image to improve accuracy. Imran et al.~\cite{imran2019depth} introduced depth coefficients, and Xu et al.~\cite{xu2019depth} suggested using surface normals as constraints. Eldesokey et al.~\cite{eldesokey2019confidence} utilized confidence to enhance dense depth prediction. Qiu et al.~\cite{Qiu_2019_CVPR} proposed fusing depth and surface normals using adaptive attention, while Chen et al.~\cite{chen2019learning} and Huynh et al.~\cite{huynh2020boosting} proposed merging appearance and geometry directly in the feature space. Cheng et al.~\cite{cheng2018depth,cheng2019learning} and Park et al.~\cite{park2020non} iteratively improved depth prediction using propagation architectures. Nevertheless, these methods are mainly developed to densify depth measurements from range sensors. Inspired by the recent depth completion method~\cite{huynh2020boosting}, we leverage its 3D point fusion architecture for the monocular depth estimation problem. 

\noindent \textbf{Attention mechanism:}
Xu et al.~\cite{xu2015show} was one of the first works utilizing attention for vision tasks. Later on, attention was implemented as spatial attention~\cite{bello2019attention,wang2018non}, channel-wise attention~\cite{hu2018squeeze,tan2019mnasnet}, and mix attention~\cite{wang2017residual} to improve classification and detection accuracy. Recent monocular depth estimation methods~\cite{huynh2020guiding,kong2019pixel,li2018deep,liu2021monocular,xu2018structured} also applied the attention mechanism. However, these attention implementations require relatively heavy computational resources.

\section{Proposed Method} \label{sec:method}

The overall structure of our FuSaNet model is shown in Figure~\ref{fig:overview}. It consists of two major components: Fusion-Net and Saliency-Net. At training time, the Fusion-Net first takes the RGB image and 3D guiding points as inputs to produce a confidence map, a depth map, and a fusion features volume. These intermediate outputs are utilized in Saliency-Net to detect a set of salient points. Both the RGB image and the salient points are fed to the network to produce the final depth prediction. However at testing time, only the RGB image is needed as input to the model.

\subsection{Fusion-Net}

\begin{figure*}[!t]
\begin{center}
  \includegraphics[width=0.99\linewidth]{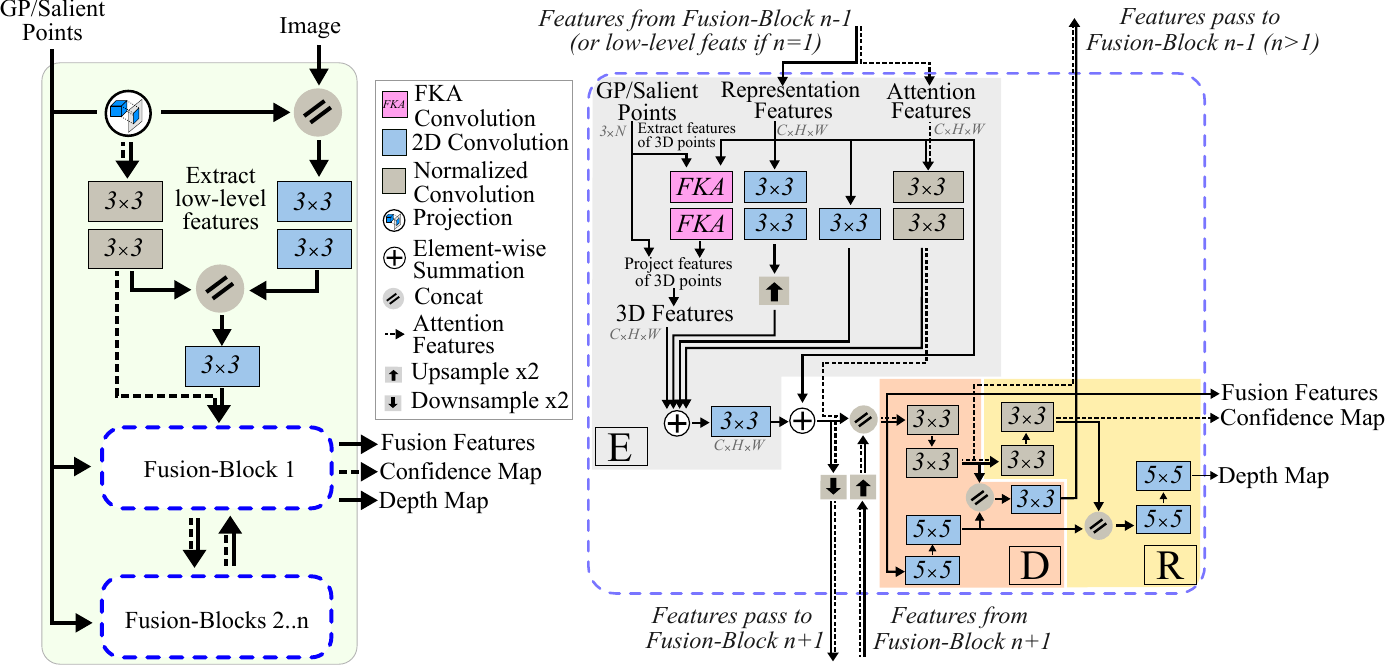}
\end{center}
  \caption{The Fusion-Net (left) consists of five Fusion-Blocks that extract and fuse 2D and 3D features at multiple-scale. The Fusion-Block (right) includes the feature fusion encoder [E], confidence predictor, decoder [D] and refinement [R] modules.}
\label{fig:architecture_overview}  \vspace{-0.45cm}
\end{figure*}

Inspired by~\cite{huynh2020boosting}, we design the Fusion-Net as a fully convolutional framework as shown in Figure~\ref{fig:architecture_overview}. Along with the RGB image, the input 3D points to the Fusion-Net can be the salient points or the 3D guiding points (GP), which the latter is only utilized during training. To generate the GP, we apply SIFT~\cite{lowe2004distinctive} to find keypoints locations from the input image and sample the GP from the ground truth depth map using these keypoint locations. However, one should notice that GP is only an initial guess of the salient points, and the network learns to detect the final points that are reliable in terms of monocular depth.

First, we obtain a sparse depth map and a sparse binary mask from the 2D locations of the input 3D points.
The RGBD image is generated by stacking the RGB image with the sparse depth map. Next, two normalized convolutional layers (nConv)~\cite{eldesokey2020uncertainty,eldesokey2019propagating} and two convolutional layers (Conv2D) are applied to the binary mask, sparse depth, and the RGBD image. Representation features of the two outputs are concatenated and passed to another Conv2D to create the low-level input features along with the attention features from the output of the previous nConvs. The Fusion-Net has five Fusion-Blocks that operate at multi-scale resolutions. Each Fusion-Block consists of a feature fusion encoder, a decoder, a refinement, and a confidence predictor module.

\noindent\textbf{Feature Fusion Encoder.} 
The feature fusion encoder is used to extract and fuse appearance, attention, and geometric features. This module takes a 3D attention tensor, a 3D representation tensor, and a set of salient points as inputs to produce the fusion features and attention volumes with the same shape as the input tensors. 

Details of this module are presented in Figure~\ref{fig:architecture_overview} (right) that consists of two 2D branches, one normalized convolution branch, one 3D branch, and one convolutional layer for feature fusion. The 2D branches are convolved at two different resolutions to learn multi-scale representations from the input volume. The normalized convolution branch learns both appearance and attention features utilizing the nConv. The 3D branch aims to extract structural features from the salient points using the feature-kernel alignment convolution (FKAConv)~\cite{boulch2020fka}. The representation features from four branches have the same shape as the input tensors that are summed together before applying a Conv2D to output a 3D tensor. Finally, a residual connection is added to avoid vanishing gradient at training time. Meanwhile, the attention features are passed to the decoder module.

\noindent\textbf{Confidence Predictor.}
The confidence predictor is a sequence of nConvs stretching from the encoder to the refinement module. Unlike the Conv2D, the nConv takes the representation features and the attention features as inputs to estimate a confidence map by propagating the attention volume weighting every pixel by its importance. 
In addition, output features from the confidence predictor are also used to guide the training of the encoder, decoder, and refinement blocks, as illustrated in Figure~\ref{fig:architecture_overview} (right).

\noindent\textbf{Decoder and Refinement Modules.} Many existing deep learning-based methods~\cite{chen2019structure,fang2020towards,Hu2018RevisitingSI,Ranftl2021,wojna2019devil} for monocular depth estimation employ complicated decoders to obtain high levels of accuracy.
However, it was shown in~\cite{huynh2020boosting,huynh2020guiding,poggi2018towards} that it is feasible to achieve competitive performance utilizing more lightweight and simplistic architectures.
Following this line of work, we design our decoder and refinement module using structure which consists of two branches: the upper branch uses several nConvs to decode learned features and predicts the confidence map, while the lower branch contains a series of Conv2Ds, that along with features from the upper one, is used to predict the depth map. Figure~\ref{fig:architecture_overview} (right) illustrates the structure of our decoder (D, the orange block) and refinement (R, the yellow block) modules. 

\begin{figure}[!t]
\begin{center}
  \includegraphics[width=0.99\linewidth]{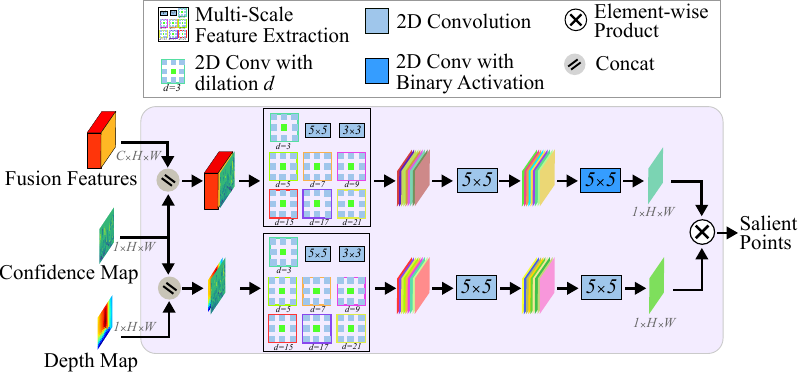}
\end{center}

  \caption{Structure of the Saliency-Net.} \vspace{-0.35cm}
\label{fig:salient_estimation}
\end{figure}

\subsection{Saliency-Net}

Figure~\ref{fig:salient_estimation} shows the overall structure of the Saliency-Net. The 3D guiding points (used only at training) and RGB image are first passed through the Fusion-Net to produce a depth map, a confidence map, and a fusion features tensor. This confidence map is utilized for guidance, and it is separately concatenated with the fusion features and the depth map before passing them to the multi-scale feature extraction layer (MFE). The MFE contains two 2D convolutional layers (3x3, 5x5) and seven 3x3 2D atrous convolutions (d = 3, 5, 7, 9, 15, 16, 21) that is used to capture features at various spatial resolutions. Finally, extracted features are fed to convolutional layers and the element-wise product to generate the salient points that serve as an input for the Fusion-Net to predict the final depth map.

\subsection{Training Loss Functions}

Our loss function consist of normalized Hessian loss, sparse loss, and depth confidence loss terms.

\noindent \textbf{Normalized Hessian loss:} 3D reconstruction from a monocular view is inherently an ill-posed problem. There are many ambiguities present and one of them is the generalized bas-relief (GBR) transformation that exists when an unknown object with Lambertian reflectance is viewed orthographically. This is a well-known property that sculpturers have exploited since ancient times to make reliefs more shallow than they actually are. To deal with this ambiguity we observe that
at each spatial location, the unit vector obtained by normalizing the three independent elements $\nabla^{2}z=\left(z_{xx},\,z_{xy},\,z_{yy}\right)$ of the Hessian of a depth map $z$, is invariant to scaling and shears along the \emph{z-}axis i.e, the GBR transformation~\cite{pedone2021learning}. We thus define the quantity $H_{z}$ based on the normalized Hessian of $z$ as:

\begin{equation}
H_{z}=\frac{\nabla^{2}z}{\left\Vert \nabla^{2}z\right\Vert +\varepsilon}\label{eq:normalize_hessian}
\end{equation}

\noindent where $\varepsilon=10^{-20}$ is a small scalar quantity introduced
to avoid divisions by zero. The quantities in $\nabla^2z$ are in practice estimated using Gaussian second derivative filters. Given a predicted depth map $\widehat{d}$
and a ground-truth depth map $d$, we can measure their dissimilarity
by calculating the root mean squared error of $H_{\widehat{d}}$ and
$H_{d}$, and formulate the following loss function:

\begin{equation}
\mathcal{L}_{H}(\widehat{d},d)=\sqrt{\frac{1}{N}\sum_{x,y}\left\Vert H_{\widehat{d}}(x,y)-H_{d}(x,y)\right\Vert ^{2}}\label{eq:loss_hessian}
\end{equation}

\noindent where $N$ denotes the total number of pixels of the depth map. Note
that since $H_{\widehat{d}}$ and $H_{d}$ are unit vectors, the term
inside the summation in (\ref{eq:loss_hessian}) represents the squared
chordal distance between two points on the unit sphere. Since the
operator $H$ is invariant to linear transformations along the optical
axis, it follows that whenever $\widehat{d}$ and $d$ are related
by a $z$-scaling or shear, their dissimilarity $\mathcal{L}_{H}(\widehat{d},d)$
will be automatically $0$, and consequently, the network will treat
scaled versions of the same depth map as an entire equivalence class.

\noindent \textbf{Sparse loss:} The $\mathcal{L}_{S}$ is defined as the ratio between the root mean square error at ground truth sparse depth and all valid depth values. This term is used to minimize the error at sparse depth positions of the estimated and ground-truth depth maps, and is defined as follow:

\begin{equation} 
\label{eq:loss_sparse}
\mathcal{L}_{S} = {\sqrt{\frac{\sum_{j=0}^{M_s}(\hat{d}_{j} - d_{j})^2}{M_s}}} /  {\sqrt{\frac{\sum_{i=0}^{M}(\hat{d}_{i} - d_{i})^2}{M}}} 
\end{equation}

\noindent where $M_s$ is the number of ground truth 3D point input and $M$ is the number of valid depth values.

\begin{table*}[t!]
\caption{\label{tab:eval_nyuv2}Evaluation on the NYU-Depth-v2 dataset. Metrics with $\downarrow$ mean lower is better and $\uparrow$ mean higher is better. Methods with $^{\ddagger}$ are trained using extra data while $^{\star\star}$ indicate the use of the whole training set.}
\centering
\small
\begin{tabular}{@{}llrrcccccc@{}}
\hline
\multicolumn{2}{c}{\textbf{Architecture}} & \textbf{\#params} & \textbf{REL$\downarrow$} & \textbf{RMSE$\downarrow$} & \(\boldsymbol{\delta_{1}}\)$\uparrow$ & \(\boldsymbol{\delta_{2}}\)$\uparrow$ & \(\boldsymbol{\delta_{3}}\)$\uparrow$ \\ \hline

Eigen \& Fergus & Eigen \& Fergus'15~\cite{eigen2015predicting}$^{\star\star}$ & 141.1M & 0.158 & 0.641 & 0.769 & 0.950 & 0.988 \\ \hline

FCRN & Laina'16~\cite{laina2016deeper}$^{\star\star}$ & 63.4M & 0.127 & 0.573 & 0.811 & 0.953 & 0.988 \\ \hline

PlaneNet & Liu'18~\cite{liu2018planenet}$^{\ddagger}$ & 47.5M & 0.142 & 0.514 & 0.812 & 0.957 & 0.989 \\ \hline

Detail Preserving-depth & Hao'18~\cite{hao2018detail} & 60.0M & 0.127 & 0.555 & 0.841 & 0.966 & 0.991 \\ \hline

Relative-depth & Lee'19~\cite{lee2019monocular} & 118.6M & 0.131 & 0.538 & 0.837 & 0.971 & 0.994 \\ \hline

DS-SIDENet & Ren'19~\cite{ren2019deep} $^{\star\star}$ & 49.8M & 0.113 & 0.501 & 0.833 & 0.968 & 0.993 \\ \hline

PAP-Net & Zhang'19~\cite{zhang2019pattern} & 95.4M & 0.121 & 0.497 & 0.846 & 0.968 & 0.994 \\ \hline

DORN & Fu'19~\cite{fu2018deep}  & 110.0M & 0.115 & 0.509 & 0.828 & 0.965 & 0.992 \\ \hline

GeoNet & Qi'19~\cite{qi2018geonet} & 67.2M & 0.128 & 0.569 & 0.834 & 0.960 & 0.990 \\ \hline

SharpNet & Ramam.'19~$^{\ddagger}$~\cite{ramamonjisoa2019sharpnet} & 80.4M & 0.139 & 0.502 & 0.836 & 0.966 & 0.993 \\ \hline

Revisited mono-depth & Hu'19~\cite{Hu2018RevisitingSI}  & 157.0M & 0.115 & 0.530 & 0.866 & 0.975 & 0.993 \\ \hline

SARPN & Chen'19~\cite{chen2019structure}  & 210.3M & 0.111 & 0.514 & 0.878 & 0.977 & 0.994 \\ \hline

VNL & Yin'19~\cite{Yin2019enforcing}  & 114.2M & 0.108 & 0.416 & 0.875 & 0.976 & 0.994 \\ \hline

BTS & Lee'20~\cite{lee2019big}  & 47.0M & 0.110 & 0.392 & 0.885 & 0.978 & 0.994 \\ \hline

DAV & Huynh'20~\cite{huynh2020guiding}  & 25.1M & 0.108 & 0.412 & 0.882 & 0.980 & 0.996 \\ \hline

AFDB-Net & Liu'21~\cite{liu2021monocular} & 139.2M & 0.113 & 0.504 & 0.878 & 0.978 & 0.995 \\ \hline

TransDepth & Yang'21~\cite{yang2021transformers}  & 311.3M & 0.106 & 0.365 & 0.900 & 0.983 & 0.996 \\ \hline

DPT & Ranftl'21$^{\ddagger}$~\cite{Ranftl2021}  & 123.9M & 0.110 & \textbf{0.357} & 0.904 & 0.988 & \textbf{0.998} \\ \hline

FuSaNet & Ours & \textbf{8.1M} & \textbf{0.104} & 0.403 & \textbf{0.915} & \textbf{0.989} & \textbf{0.998} \\ \hline

\end{tabular} \vspace{-0.25cm}
\end{table*}

\noindent \textbf{Depth Confidence loss:} This loss term is defined as:

\begin{equation} 
\label{eq:loss_depth_confidence}
\mathcal{L}_{DC} = \mathcal{L}_{log} + \mu \mathcal{L}_{grad} + \theta \mathcal{L}_{norm} - \psi\frac{1}{p}\mathcal{L}_{C}
\end{equation}

\noindent where $\mathcal{L}_{log}$ is a variation of the $L_{1}$ norm that minimizes error on the depth pixels, $\mathcal{L}_{grad}$ optimizes the error on edge structures, and $\mathcal{L}_{norm}$ penalizes angular error between the ground truth and predicted normal surfaces. These loss terms were introduced by Hu et al.~\cite{Hu2018RevisitingSI} and widely adopted by state-of-art monocular depth estimation methods~\cite{chen2019structure,huynh2020guiding}. We adopt the confidence loss proposed by~\cite{eldesokey2019confidence} where $p$ is the training epoch and $\mathcal{L}_{C} = C - \mathcal{L}_{log}C$ where $C$ is the predicted confidence map. The full loss function that we utilize is

\begin{equation} 
\label{eq:loss_total}
\mathcal{L}_{total} = \sum_{i=1}^{n=5} \gamma^{i} (\beta \mathcal{L}^{i}_{DC} + \phi \mathcal{L}^{i}_{S} + \lambda \mathcal{L}^{i}_{H} )
\end{equation}

\noindent where $n$ is the number of resolution scales and $\gamma^{i} \in \mathcal{R}^{+}$ is the loss weight at scale $i$; $\beta, \phi, \lambda \in \mathcal{R}^{+}$ are weight loss coefficients. Subsection~\ref{sec:implementation_detail} describes in detail how the network is trained using these loss functions.

\section{Experiments} \label{sec:experiment}

In this section, we first describe the datasets and evaluation metrics that are used in our experiments followed by the implementation details of our model. The rest of this section presents a comparison with the state-of-the-art, ablation studies, and qualitative results.

\subsection{Dataset and Metrics}
We evaluate our proposed model on three datasets: NYU-Depth-v2~\cite{silberman2012indoor}, KITTI~\cite{geiger2013vision}, and SUN-RGBD~\cite{janoch2013category,song2015sun,xiao2013sun3d}. NYU-Depth-v2 includes 120K RBG-D images captured from 464 indoor scenes, and we sample 50K images from the entire dataset to train and 654 test images of 215 scenes to test our model. To evaluate on KITTI outdoor driving dataset, we use the standard Eigen split~\cite{eigen2015predicting,eigen2014depth} for training (39K images) and testing (697 images). SUN-RGBD contains more than 10K images from a variety of indoor scenarios. We test our pre-trained model on 5050 images from its test set without any fine-tuning for this dataset. We follow the previous methods~\cite{eigen2015predicting,eigen2014depth} for evaluation by using the following metrics: mean relative error (REL), root mean square error (RMSE), thresholded accuracy($\delta_{i}$), scale-invariant mean square error (SI), mean absolute error (iMAE) and root mean square error (iRMSE) of the inverse depth values.

\subsection{Implementation Details} \label{sec:implementation_detail}

\begin{table}[b!]
\vspace{-0.25cm}
\caption{\label{tab:eval_kitti}Evaluation on the KITTI dataset. Metrics with $\downarrow$ mean lower is better and $\uparrow$ mean higher is better.}
\centering
\small
\adjustbox{width=\columnwidth}{\begin{tabular}{@{}lrccccc@{}}
\hline
\textbf{Method} &\textbf{\#params} & \textbf{REL$\downarrow$} & \textbf{RMSE$\downarrow$} & \(\boldsymbol{\delta_{1}}\)$\uparrow$ & \(\boldsymbol{\delta_{2}}\)$\uparrow$ & \(\boldsymbol{\delta_{3}}\)$\uparrow$ \\ \hline

DORN~\cite{fu2018deep}  & 110.0M & 0.072 & 2.626 & 0.932 & 0.984 & 0.994 \\ \hline

AFDB-Net~\cite{liu2021monocular} & 139.2M & 0.071 & 2.848 & 0.933 & 0.983 & 0.995 \\ \hline

VNL~\cite{Yin2019enforcing}  & 114.2M & 0.072 & 3.258 & 0.938 & 0.990 & 0.998 \\ \hline

BTS~\cite{lee2019big}  & 112.8M & \textbf{0.059} & 2.756 & 0.956 & 0.993 & 0.998 \\ \hline

PGA-Net~\cite{xu2020probabilistic}  & 168.3M & 0.063 & 2.634 & 0.952 & 0.992 & 0.998 \\ \hline

TransDepth~\cite{yang2021transformers}  & 311.3M & 0.064 & 2.755 & 0.956 & 0.994 & \textbf{0.999} \\ \hline

DPT~\cite{Ranftl2021}  & 123.9M & 0.062 & 2.573 & 0.959 & \textbf{0.995} & \textbf{0.999} \\ \hline

FuSaNet (Ours) & \textbf{8.1M} & \textbf{0.059} & \textbf{2.487} & \textbf{0.964} & \textbf{0.995} & \textbf{0.999} \\ \hline

\end{tabular}}
\end{table}

The proposed model is trained for 150 epochs on a single TITAN RTX using batch size of 32, and the Adam optimizer~\cite{kingma2014adam} with $(\beta_1, \beta_2, \epsilon) = (0.9, 0.999, 10^{-8})$. We use all of the loss terms in Eq.~\ref{eq:loss_total} to train the NYU-Depth-v2 dataset while using the $L_{log}$ and the $L_S$ for training KITTI. The initial learning rate is $7*10^{-4}$, but from epoch 10 the learning is reduced by $5\%$ per $5$ epochs. We set the number of scales $n$ in Eq.~\ref{eq:loss_total} to 5, weight loss coefficients $\mu, \theta, \beta, \psi$ to $1.0$; $\lambda$ to $0.01$; $\phi$ to $5.0$ for NYU-Depth-v2 and $20.0$ for KITTI. The scale weight losses $\gamma^1, \gamma^2, \gamma^3, \gamma^4, \gamma^5$ to $1.0, 0.75, 0.5, 0.25$ and $0.125$ respectively. During training, we augment the input RGB and ground truth depth images using random rotation ([-5.0, +5.0] degrees), horizontal flip, rectangular window droppings, and colorization (RGB only). We train the network using 3D points sampled from ground truth depth maps at keypoint locations and for some random iteration feeding zero points during training. At testing time, the model only takes the RGB image as input.

\subsection{Performance Analysis}

Tables~\ref{tab:eval_nyuv2},~\ref{tab:eval_kitti} and~\ref{tab:eval_generalization} present the quantitative comparison and the number of model parameters for our method and state-of-the-art approaches. The results show that, the proposed method, while being the smallest model, achieves comparable figures against the baselines.

\begin{figure*}[!t]
\begin{center}
    \includegraphics[width=0.93\linewidth]{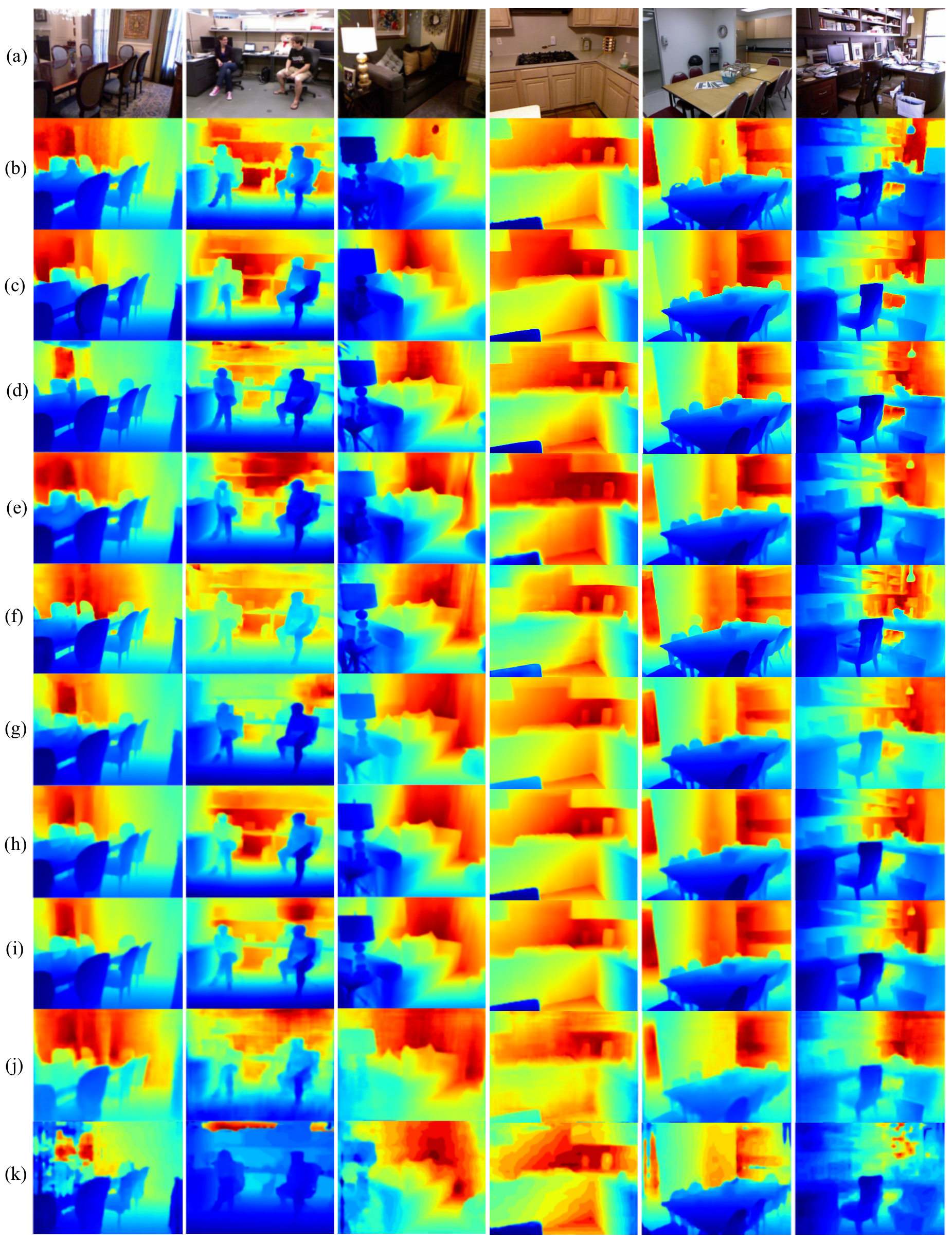}
\end{center}

  \caption{Examples from the NYU-Depth-v2 test set.
  (a) Input images, (b) ground truth depth. Results from 
  (c) FuSaNet,
  (d) DPT~\cite{Ranftl2021},
  (e) DAV~\cite{huynh2020guiding},
  (f) BTS~\cite{lee2019big},
  (g) VNL~\cite{Yin2019enforcing},
  (h) SARPN~\cite{chen2019structure},
  (i) Hu et al.~\cite{Hu2018RevisitingSI},
  (j) SharpNet~\cite{ramamonjisoa2019sharpnet},
  and (k) DORN~\cite{fu2018deep}.
  } %
\label{fig:qualitative_nyu} %
\end{figure*}

In case of NYU-Depth-v2, the best approaches including TransDepth~\cite{yang2021transformers}, SARPN~\cite{chen2019structure}, DPT~\cite{Ranftl2021}, VNL~\cite{Yin2019enforcing}, BTS~\cite{lee2019big}, and DAV~\cite{huynh2020guiding} use 38.4, 25.9, 15.3, 14.1, 5.8 and 3.1 times more parameters in contrast to ours, respectively. 
Compared to DPT~\cite{Ranftl2021}, our depth maps have difficulties in tiny detailed structures such as leaves, legs of the chair. Nevertheless, our model produces high-quality depth maps compared to state-of-the-art approaches that employ large network~\cite{chen2019structure,Hu2018RevisitingSI,huynh2020guiding,lee2019big,ramamonjisoa2019sharpnet,Ranftl2021,Yin2019enforcing} and train with a large amount of extra data~\cite{ramamonjisoa2019sharpnet,Ranftl2021}, as shown in Figure~\ref{fig:qualitative_nyu}. Furthermore, the proposed method model performs well in uniform regions and large furniture like bookshelves, tables.

\begin{figure*}[!t]
\begin{center}
    \includegraphics[width=0.93\linewidth]{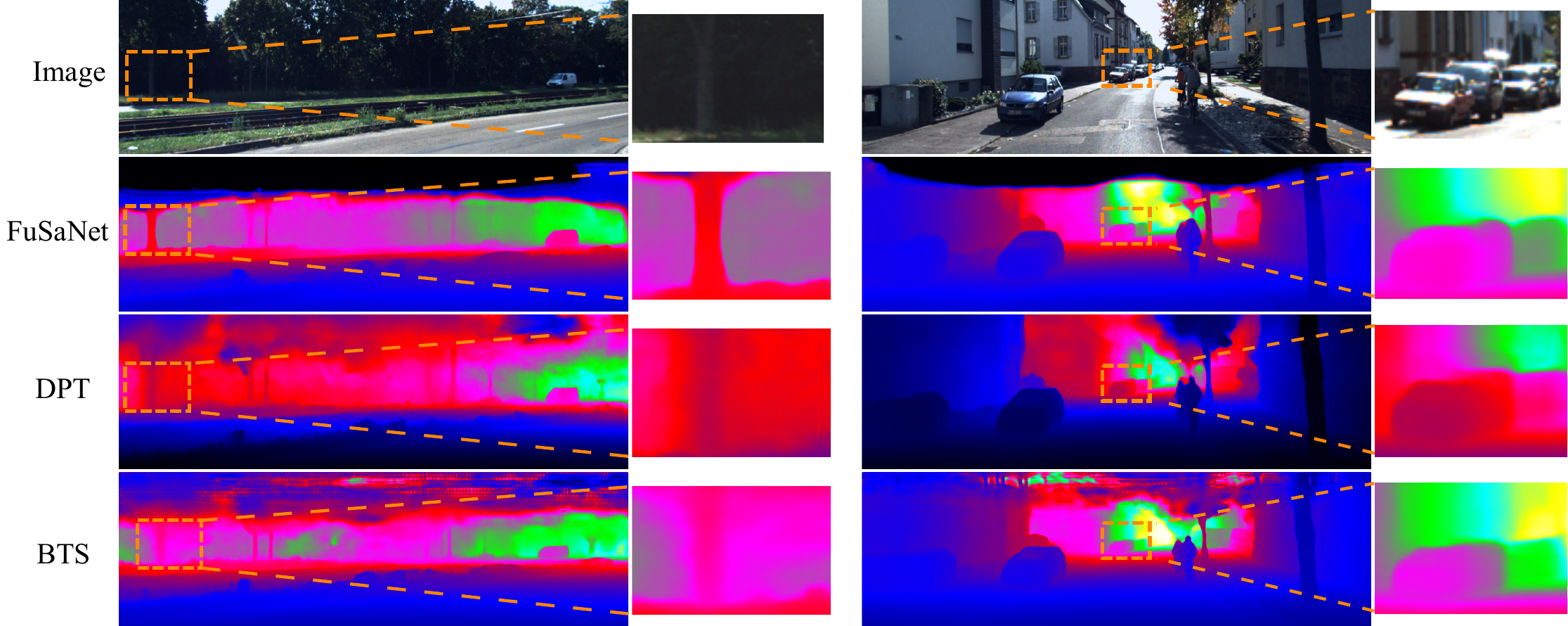}
\end{center}
  \vspace{-0.05cm}
  \caption{Comparison with BTS~\cite{lee2019big} and DPT~\cite{Ranftl2021} approaches on  the KITTI dataset. Each example contains an image or a predicted depth map (left) with a zoom-in view (right) for visualization.
  } %
\label{fig:qualitative_kitti} %
\end{figure*}

For KITTI, our approach performs on par with state-of-the-art methods while being (at least 13.5 times) more compact in terms of the number of parameters. As shown in Figure~\ref{fig:qualitative_kitti}, our model yields high quality depth predictions, especially at object boundaries and contours.

\begin{figure*}[!t]
\begin{center}
  \includegraphics[width=0.99\linewidth]{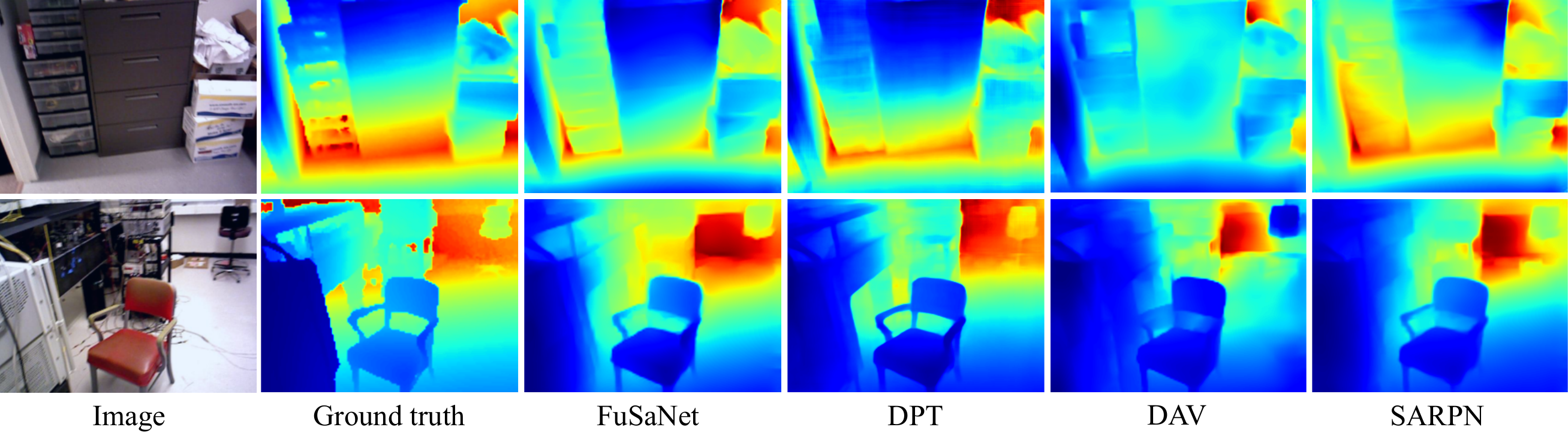}
\end{center}
  \vspace{-0.05cm}
  \caption{Cross-dataset evaluation on SUN RGB-D dataset. SARPN~\cite{chen2019structure}, DAV~\cite{huynh2020guiding} and FuSaNet models were trained on NYU-Depth-v2 while DPT~\cite{Ranftl2021} was trained on \textit{MIX 6}~\cite{Ranftl2021} before fine-tuning on NYU-Depth-v2.
  } 
\label{fig:qualitative_sunrgbd} %
\end{figure*}

Moreover, to assess the generalizability of our network, we perform a cross-dataset evaluation, where we train the model using NYU-Depth-v2 and test with SUN-RGBD without any fine-tuning. We also evaluate the methods from \cite{chen2019structure,Hu2018RevisitingSI,huynh2020guiding,Ranftl2021} and present the results in Table~\ref{tab:eval_generalization} and Figure~\ref{fig:qualitative_sunrgbd}. As can be seen our model performs favourably compared to baseline approaches.

\begin{table}[t!]
\caption{\label{tab:eval_generalization}Cross-dataset evaluation with training on NYU-Depth-v2 and testing on SUN-RGBD.}
\centering
\small
\adjustbox{width=\columnwidth}{\begin{tabular}{@{}lrccccc@{}}
\hline
\textbf{Models} & \textbf{\#params} & \textbf{REL} & \textbf{sqREL} & \textbf{SI} & \textbf{iMAE} & \textbf{iRMSE} \\ \hline
Hu et al.~\cite{Hu2018RevisitingSI} & 157.0M & 0.245 & 0.389 & 0.031 & 0.108 & 0.087 \\ \hline

SARPN~\cite{chen2019structure} & 210.3M & 0.243 & 0.393 & 0.031 & \textbf{0.102} & 0.069 \\ \hline
 
DAV~\cite{huynh2020guiding} & 25.1M & 0.238 & 0.387 & 0.030 & 0.104 & 0.075 \\ \hline
 
DPT~\cite{Ranftl2021} & 123.9M & 0.230 & \textbf{0.341} & \textbf{0.029} & 0.103 & \textbf{0.068} \\ \hline
 
FuSaNet (Ours) & \textbf{8.1M} & \textbf{0.225} & 0.350 & \textbf{0.029} & \textbf{0.102} & 0.070 \\ \hline
\end{tabular}} \vspace{-0.45cm}
\end{table}

\begin{table}[b!]
\vspace{-0.25cm}
\caption{\label{tab:ablation_salient_net}Ablation studies of models without and with the Saliency-Net on the NYU-Depth-v2 dataset. Frame rate \textit{(fps)} is measured using one GTX-1080 GPU.}
\centering
\small

\adjustbox{width=\columnwidth}{\begin{tabular}{@{}lccccc@{}}
\hline
\textbf{Training} & \textbf{Frame rate$\uparrow$} & \textbf{REL$\downarrow$} & \textbf{RMSE$\downarrow$} & \(\boldsymbol{\delta_1}\)$\uparrow$ & \(\boldsymbol{\delta_2}\)$\uparrow$ \\ \hline
w/o Saliency-Net & \textbf{327} & 0.122 & 0.494 & 0.857 & 0.971  \\ \hline

\textbf{w/ Saliency-Net} & 172 & \textbf{0.104} & \textbf{0.403} & \textbf{0.915} & \textbf{0.989}  \\ \hline
 
\end{tabular}}
\end{table}

\subsection{Ablation Studies}

\noindent \textbf{Effectiveness of the Saliency-Net:}\quad We implement models with and without the Saliency-Net to assess its effect to the performance. As presented in Table~\ref{tab:ablation_salient_net}, applying the proposed module significantly improves the accuracy while increasing the runtime.
 
\noindent \textbf{3D guiding points at training time:}\quad We experiment with three different schemes to study the impact of using the 3D guiding points at training time and report the results in Table~\ref{tab:ablation_3d_pts}. \textit{FuSaNet1} is the model trained with only RGB images, while \textit{FuSaNet2} also utilized the 3D guiding points as the input. The results show that \textit{FuSaNet2} significantly outperforms \textit{FuSaNet1} since the model trained with 3D guiding points is better at exploiting monocular priors from RGB images even when no points are fed to the network during testing. 
Moreover, we experiment with \textit{FuSaNet3} by adopting a similar training procedure to \textit{FuSaNet2}, except that we feed zero points to the network for some random iterations. This training procedure further robustifies our model to the absence of 3D points, as shown in Table~\ref{tab:ablation_3d_pts}.

\begin{figure*}[!t]
\begin{center}
    \includegraphics[width=0.99\linewidth]{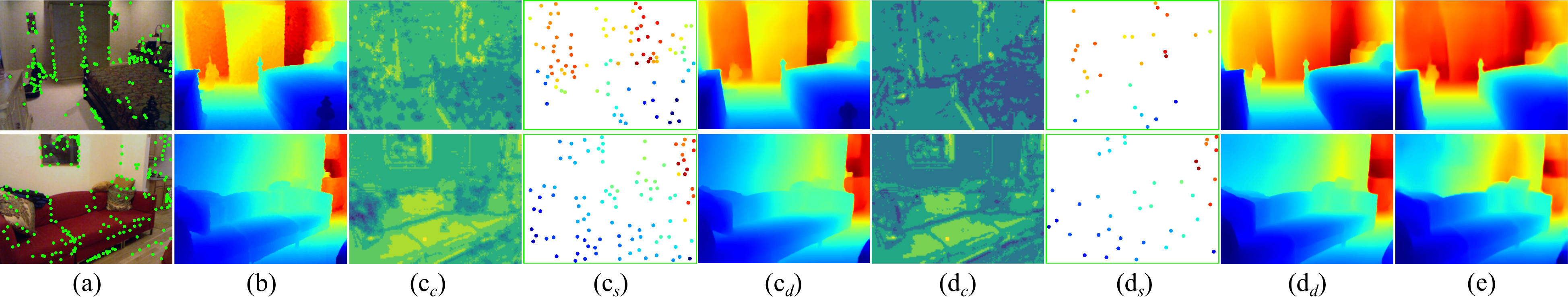}
\end{center}
  \vspace{-0.1cm}
  \caption{Examples from the NYU-Depth-v2 test set. 
  From left to right:
  (a) RGB images and 3D guiding points inputs; 
  (b) ground truth depth; 
  (c$_{c}$) confidence map, (c$_{s}$) salient points, and (c$_{d}$) final predicted depth
  of model using the RGB image and 500 guiding points as inputs; 
  (d$_{c}$), (d$_{s}$), (d$_{d}$) similar results of model using only the RGB image as input;
  and (e) predicted depth of model without the confidence predictor and using only the RGB image as input.
  (Points are enhanced for visualization)
  } %
\label{fig:confidence_maps} %
\end{figure*}

\noindent \textbf{Confidence predictor (CP):}\quad We conduct experiments with and without the confidence predictor to analyze how this module affects the performance by following the training scheme of \textit{FuSaNet3} in Table~\ref{tab:ablation_3d_pts}. 
As shown in Table~\ref{tab:ablation_confidence_predictor}, the results clearly demonstrate the benefit of the CP block. Furthermore, we observe that the CP learns to highlight important locations from RGB images either with or without input 3D points at the inference time, as presented in the confidence maps (c$_c$, d$_c$) and salient points (c$_s$, d$_s$) in Figure~\ref{fig:confidence_maps}. This, in turn, helps the proposed network to produce high-quality depth maps (d$_{d}$, Figure~\ref{fig:confidence_maps}) with less distortion than the model without the CP module (e, Figure~\ref{fig:confidence_maps}).

\noindent \textbf{Training losses:}\quad
We also study the impact of different loss terms by training our method with various settings and report the results in Table~\ref{tab:ablation_losses}.  Models that incorporated the normalized Hessian loss show clear improvements in all metrics.

\begin{table}[t!]
\caption{\label{tab:ablation_3d_pts}Ablation studies of models using the 3D guiding points (GP) during training on NYU-Depth-v2. FuSaNet2 is trained using the RGB image and GP while FuSaNet1 uses only RGB information. FuSaNet3 is also trained with GP but for some random iteration no points are fed to the network at training time.}
\centering
\small
\begin{tabular}{lccccc}
\hline
\textbf{Training} & \textbf{REL$\downarrow$} & \textbf{RMSE$\downarrow$} & \(\boldsymbol{\delta_1}\)$\uparrow$ & \(\boldsymbol{\delta_2}\)$\uparrow$ & \(\boldsymbol{\delta_3}\)$\uparrow$ \\ \hline
FuSaNet1 & 0.119 & 0.474 & 0.857 & 0.978 & 0.994 \\ \hline

FuSaNet2 & 0.105 & 0.416 & 0.912 & 0.988 & 0.996 \\ \hline

\textbf{FuSaNet3} & \textbf{0.104} & \textbf{0.403} & \textbf{0.915} & \textbf{0.989} & \textbf{0.998} \\ \hline
\end{tabular} \vspace{-0.45cm}
\end{table}

\begin{table}[b!]
\caption{\label{tab:ablation_confidence_predictor}Ablation studies of models without and with the confidence predictor (CP) on NYU-Depth-v2.}
\centering
\small
\begin{tabular}{lccccc}
\hline
\textbf{Training} & \textbf{REL$\downarrow$} & \textbf{RMSE$\downarrow$} & \(\boldsymbol{\delta_1}\)$\uparrow$ & \(\boldsymbol{\delta_2}\)$\uparrow$ & \(\boldsymbol{\delta_3}\)$\uparrow$ \\ \hline
w/o CP & 0.126 & 0.530 & 0.843 & 0.967 & 0.992 \\ \hline

\textbf{w/ CP} & \textbf{0.104} & \textbf{0.403} & \textbf{0.915} & \textbf{0.989} & \textbf{0.998} \\ \hline
\end{tabular}
\end{table}

\begin{table}[b!]

\caption{\label{tab:ablation_losses}Ablation studies of different loss terms on the NYU-Depth-v2 dataset. Results are obtained from one iteration.}
\centering
\small
\adjustbox{width=\columnwidth}{\begin{tabular}{@{}lccccc@{}}
\hline
{\textbf{Training}} & \textbf{REL$\downarrow$} & \textbf{RMSE$\downarrow$} & \(\boldsymbol{\delta_{1}}\)$\uparrow$ & \(\boldsymbol{\delta_{2}}\)$\uparrow$ & \(\boldsymbol{\delta_{3}}\)$\uparrow$ \\ \hline

FuSaNet + $L_{DC}$ & 0.119 & 0.452 & 0.871 & 0.976 & 0.994 \\ \hline

FuSaNet + $L_{DC}$ + $L_{S}$ & 0.116 & 0.451 & 0.872 & 0.977 & 0.994 \\ \hline

FuSaNet + $L_{DC}$ + $L_{H}$ & 0.111 & 0.429 & 0.883 & 0.980 & 0.996 \\ \hline

\textbf{FuSaNet + $\boldsymbol{L_{DC}}$ + $\boldsymbol{L_{S}}$ + $\boldsymbol{L_{H}}$} & \textbf{0.104} & \textbf{0.403} & \textbf{0.915} & \textbf{0.989} & \textbf{0.998} \\ \hline

\end{tabular}}
\end{table}

\section{Conclusion}
This paper proposed a novel saliency-based self-attention mechanism and a normalized Hessian loss function to estimate high-quality depth prediction for indoor and outdoor environments. The proposed method achieves competitive performance while being at least three times more compact than state-of-the-art approaches. Our work provides a potential approach toward optimizing accuracy and network size for dense depth estimation without the need for using active depth sensors or multiple view geometry.

{\small
\bibliographystyle{ieee_fullname}
\bibliography{ms}
}

\end{document}